\documentclass[a4paper]{article}
\usepackage{iwslt15,amssymb,amsmath,epsfig}
\usepackage{booktabs}
\def\reg{{\rm\ooalign{\hfil
     \raise.07ex\hbox{\scriptsize R}\hfil\crcr\mathhexbox20D}}}

\title{Neural Machine Translation Training in a Multi-Domain Scenario}

\name{Hassan Sajjad, \hspace{1mm} Nadir Durrani, \hspace{1mm} Fahim Dalvi, \hspace{1mm} $^*$Yonatan Belinkov, \hspace{1mm} Stephan Vogel
\address{Qatar Computing Research Institute -- HBKU \\
$^*$MIT Computer Science and Artificial Intelligence Laboratory \\
{\small \tt \{hsajjad,ndurrani,faimaduddin,svogel\}@qf.org.qa, $^*${\tt belinkov@mit.edu}
}
}
}

\address{
}
\begin{document}
\maketitle
\begin{abstract}

In this paper, we explore alternative ways to train a neural machine translation system in a multi-domain scenario. We investigate data concatenation (with fine tuning), model stacking (multi-level fine tuning), data selection and multi-model ensemble. 
Our findings 
show that the best translation quality can be achieved by building an initial system on a concatenation of available out-of-domain data and then fine-tuning it on in-domain data. Model stacking works best when training begins with the furthest out-of-domain data and the model is incrementally fine-tuned with the next furthest domain and so on.
Data selection did not give the best results, but can be considered as a decent compromise between training time and translation quality. 
A weighted ensemble of different individual models performed better than data selection. It is beneficial in a scenario when there is no time for fine-tuning an already trained model.

\end{abstract}

\section{Introduction}

Neural machine translation (NMT) systems 
are sensitive to the data they are trained on. 
The available parallel corpora come from various genres and have different stylistic variations and 
semantic ambiguities. 
While 
such data is often beneficial for a general purpose machine translation system, a problem arises when building systems for specific domains such as lectures \cite{guzman-sajjad-etal:iwslt13,cettolo2014report}, patents \cite{Fujii10overviewof} or medical text \cite{bojar-EtAl:2014:W14-33}, where either the in-domain bilingual text does not exist or is available in small 
quantities.

Domain adaptation aims to preserve the identity of the in-domain data while exploiting the out-of-domain data in favor of the in-domain data and avoiding possible drift towards out-of-domain jargon and style. 
%
The most commonly used approach to train a domain-specific neural MT system is to fine-tune an existing model (trained on generic data) with the new domain
\cite{luong-manning:iwslt15,FreitagA16,ServanCS16,ChuDK17} or 
to add domain-aware tags in building a concatenated system \cite{KobusCS16}. 
\cite{wees:da:emnlp17} proposed a gradual fine-tuning method that starts training with complete in- and out-of-domain data and gradually reduces the out-of-domain data for next epochs. Other approaches that have been recently proposed for domain adaptation of neural machine translation are instance weighting \cite{wang:da:emnlp17,chen:da:wnmt17} and data selection \cite{wang:da:acl17}. 


%

In this paper we explore NMT in a multi-domain scenario. 
Considering a small in-domain corpus and a number of out-of-domain corpora, we target questions like: 

\begin{itemize}

\item What are the different ways to combine multiple domains during a training process?
\item What is the best strategy to build an optimal in-domain system?
\item Which training strategy results in a robust system?
\item Which strategy should be used to build a decent in-domain system given limited time?
\end{itemize}
%
To answer these, we try the following approaches: \textbf{i) data concatenation:} train a system by concatenating all the available in-domain and out-of-domain data; \textbf{ii) model stacking:} build NMT in an online fashion starting from the most distant domain, fine-tune on the closer domain and finish by fine-tuning the model on the in-domain data; \textbf{iii) data selection:} select a certain percentage of the available out-of-domain corpora that is closest to the in-domain data and use it for training the system; \textbf{iv) multi-model ensemble:} separately train models for each available domain and combine them during decoding using balanced or weighted averaging. 
We experiment with Arabic-English and German-English language pairs. Our results demonstrate the following findings: 
\begin{itemize}
\item A concatenated system fine-tuned on the in-domain data achieves the most optimal in-domain system.
\item Model stacking works best when 
starting from the furthest domain, fine-tuning on closer domains and then 
finally fine-tuning on the in-domain data.
\item A concatenated system on all available data results in the most robust system.
\item Data selection 
gives a decent trade-off between translation quality and training time.
\item Weighted ensemble is helpful when several individual models have been already 
trained 
and there is no time for retraining/fine-tuning.
\end{itemize}

The paper is organized as follows:
Section \ref{sec:approaches} describes the adaptation approaches explored in this work. We present experimental design in Section \ref{sec:experiments}. Section \ref{sec:results} summarizes the results and Section \ref{sec:conclusion} concludes. 





\section{Approaches}
\label{sec:approaches}
\begin{figure} 
	\centering
	\includegraphics[width=\linewidth]{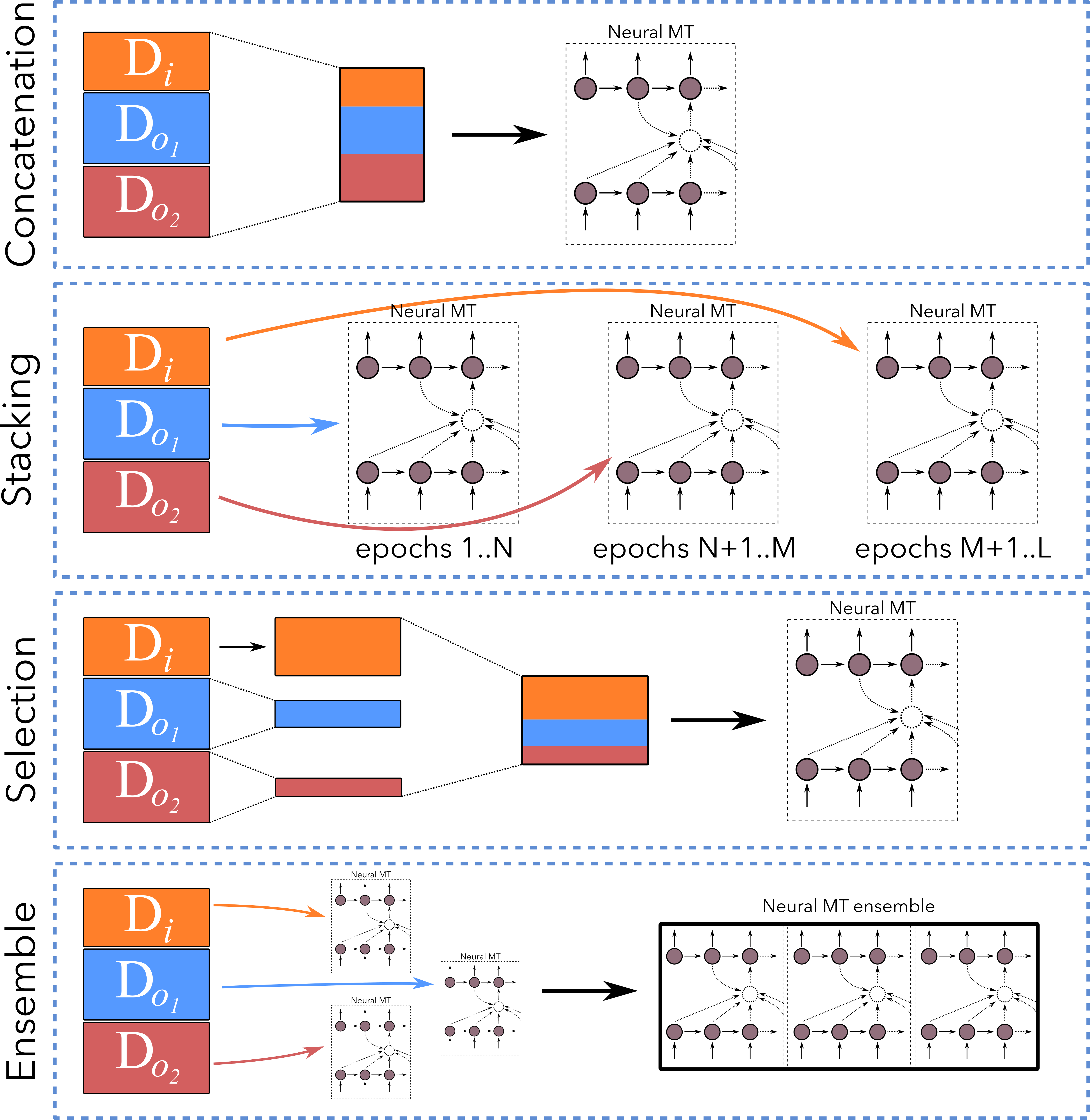}
	\caption{Multi-domain training approaches}
	\label{fig:data-approaches}
\end{figure}

Consider an in-domain data D$_i$ and a set of out-of-domain data D$_o$ = {D$_{o_1}$, D$_{o_2}$, ..D$_{o_n}$}. We explore several methods to benefit from the available data with an aim to optimize translation quality on the in-domain data. Specifically, we try data concatenation, model stacking, data selection and ensemble. Figure \ref{fig:data-approaches} presents them graphically. In the following, we describe each approach briefly.

\subsection{Concatenation}

A na\"ive yet commonly used method when training both statistical \cite{williams-EtAl:2016:WMT}\footnote{State-of-the-art baselines are trained on plain concatenation of the data with MT feature functions (such as Language Model) skewed towards in-domain data, through interpolation.} and neural machine translation systems \cite{sennrich-haddow-birch:2016:WMT} is to simply concatenate all the bilingual parallel data before training the system. During training an in-domain validation set is used to guide the training loss. 
The resulting system has an advantage of seeing a mix of all available data at every time interval, and is thus robust to handle heterogeneous test data. 


\subsection{Fine Tuning and Model Stacking}

Neural machine translation follows an online training strategy. It sees only a small portion of the data in every training step and estimates the value of network parameters based on that portion. Previous work has exploited this strategy in the context of domain adaptation. \cite{luong-manning:iwslt15} trained an initial model on an out-of-domain data and later extended the training on in-domain data. In this way the final model parameters are tuned towards the in-domain data. The approach is referred as \emph{fine-tuning} later on.

Since in this work we deal with several domains, we propose a stacking method that uses 
multi-level
fine-tuning to train a system. 
%
%
Figure \ref{fig:data-approaches} 
(second row) shows the complete procedure: first, 
the
model is trained on the out-of-domain data D$_{o_1}$ for $N$ epochs; training is resumed from $N+1$-th epoch 
to the $M$-th epoch
but using the 
next 
available out-of-domain data D$_{o_2}$; repeat the process till all of the available out-of-domain corpora have been used; in the last step, resume training on the in-domain data D$_i$ for a few 
epochs.
%
The resulting model has seen all of the available data as in the case of 
the 
data concatenation approach. However, here the system learns from the data domain by domain. We call this technique \emph{model stacking}. 

The model stacking and fine-tuning approaches 
have the 
advantage of seeing the in-domain data in the end 
of training, 
thus making the system parameters more optimized for the in-domain data. They also provide flexibility in extending an existing model to any new domain without having to retrain the complete system again on the available corpora.


\subsection{Data Selection}


Building a model, whether concatenated or stacked, on all the available data is computationally expensive. 
%
An alternative approach is \emph{data selection}, where we select a part of the out-of-domain data which is close to the in-domain data for training. The intuition here is two 
fold:
i) 
the 
out-of-domain data is huge and takes a lot of time to train on, and ii) not all parts of the out-of-domain data are beneficial for the in-domain data. 
Training only on a selected part of the out-of-domain data reduces the training time significantly 
while at the same time creating
a model closer to the in-domain.  

In this work, we use 
the
modified Moore-Lewis \cite{Axelrod_2011_emnlp} for data selection.
It trains in- and out-of-domain n-gram models and then ranks sequences in the out-of-domain data based on cross-entropy difference. The out-of-domain sentences below a certain threshold 
are selected for training. Since we are dealing with several out-of-domain corpora, we apply data selection separately on each of them and build a concatenated system using 
in-domain
plus selected out-of-domain data as shown in Figure \ref{fig:data-approaches}.
Data selection significantly reduces data size thus 
improving training time for NMT. However, finding the optimal threshold to filter data is a cumbersome process. Data selection using joint neural networks has been explored in \cite{durraniEtAl:MT-Summit2015}. We explore data selection as an alternative to the above mentioned techniques.

\subsection{Multi-domain Ensemble}

Out-of-domain data is generally available in larger quantity. Training a concatenated system whenever a new in-domain becomes available 
is expensive in terms of both time and computation. An alternative to fine-tuning the system with new in-domain is to do ensemble of the new model with the existing model.
The ensemble approach brings the flexibility to use them
during decoding without a need of retraining and fine-tuning. 

Consider $N$ models that we would like to use to generate translations. For each decoding step, we use the scores over the vocabulary from each of these $N$ models and combine them by averaging. We then use these averaged scores to choose the output word(s) for each hypothesis in our beam. The intuition is to combine the knowledge of the $N$ models to generate a translation. We refer to this approach as \emph{balanced ensemble} later on. Since 
here
we deal with several different domains, averaging scores of all the models equally may not result in optimum performance. We explore a variation of balanced ensemble called \emph{weighted ensemble} that performs a weighted average of these scores, where the weights can be pre-defined or learned on a development set.
 
Balanced ensemble using several models of a single training run saved at different iterations has shown to improve performance by 1-2 BLEU points \cite{sennrich-haddow-birch:2016:WMT}. Here our goal is not to improve the best system but to benefit from individual models built using several domains during a single decoding process. We experiment with both balanced and weighted ensemble under 
the
multi-domain condition only.\footnote{Weighted fusion of Neural Networks trained on different domains has been explored in \cite{durrani-EtAl:2016:COLING} for phrase-based SMT. Weighted training for Neural Network Models has been proposed in \cite{joty-etAL:2015:EMNLP}.} 

\section{Experimental Design}
\label{sec:experiments}

\subsection{Data}
We experiment with Arabic-English and German-English language pairs 
using 
the 
WIT$^3$ TED corpus 
\cite{cettolol:SeMaT:2016} made available for IWSLT 2016 as our in-domain data. For Arabic-English, we take the UN corpus \cite{ZiemskiJP16}  and the OPUS corpus \cite{LISON16.947} as out-of-domain corpora.
For German-English, we use the Europarl (EP), and the Common Crawl (CC) corpora made available for the {\it $1^{st}$} Conference on Statistical Machine Translation\footnote{http://www.statmt.org/wmt16/translation-task.html} as out-of-domain corpus. We tokenize Arabic, German and English using the default \emph{Moses} tokenizer. We did not do morphological segmentation of Arabic. Instead we apply sub-word based segmentation \cite{sennrich-haddow-birch:2016:P16-12} that implicitly segment as part of the compression process.\footnote{\cite{sajjad-etal:2017:ACLShort} showed that using BPE performs comparable to 
morphological tokenization \cite{abdelali-EtAl:2016:N16-3} in Arabic-English machine translation.} 
Table \ref{tab:corpusstats} shows the data statistics after running the Moses tokenizer. 

We use a concatenation of dev2010 and tst2010 sets for validation during 
training. Test sets tst2011 and tst2012 served as development sets 
to find the best model for fine-tuning and tst2013 and tst2014 are used for evaluation. We use BLEU \cite{Papineni:Roukos:Ward:Zhu:2002} to measure performance. 

\begin{table}
\centering
\begin{tabular}{lrrr}
\toprule

\multicolumn{4}{c}{\bf Arabic-English} \\
Corpus & Sentences & Tok${_{ar}}$ & Tok${_{en}}$ \\
\midrule
TED & 229k & 3.7M & 4.7M  \\
UN & 18.3M & 433M & 494M   \\
OPUS & 22.4M & 139M & 195M \\
\midrule
\multicolumn{4}{c}{\bf German-English} \\
Corpus & Sentences & Tok${_{de}}$ & Tok${_{en}}$ \\
\midrule
TED & 209K & 4M & 4.2M \\
EP  & 1.9M & 51M & 53M \\
CC  & 2.3M & 55M & 59M \\
\bottomrule
\end{tabular}
\caption{\label{tab:corpusstats} Statistics of the Arabic-English and German-English training corpora in terms of Sentences and Tokens. EP = Europarl, CC = Common Crawl, UN = United Nations.}
\end{table}

\subsection{System Settings}

We use the Nematus tool \cite{sennrich-EtAl:2017:EACLDemo} to train a 2-layered LSTM encoder-decoder with attention \cite{bahdanau:ICLR:2015}. We use the default settings: embedding layer size: 512, hidden layer size: 1000. We limit the vocabulary to 50k words 
using BPE \cite{sennrich-haddow-birch:2016:P16-12} with 50,000 operations. 

\section{Results}
\label{sec:results}

\begin{figure*}[t]
	\centering
	\includegraphics[width=\textwidth]{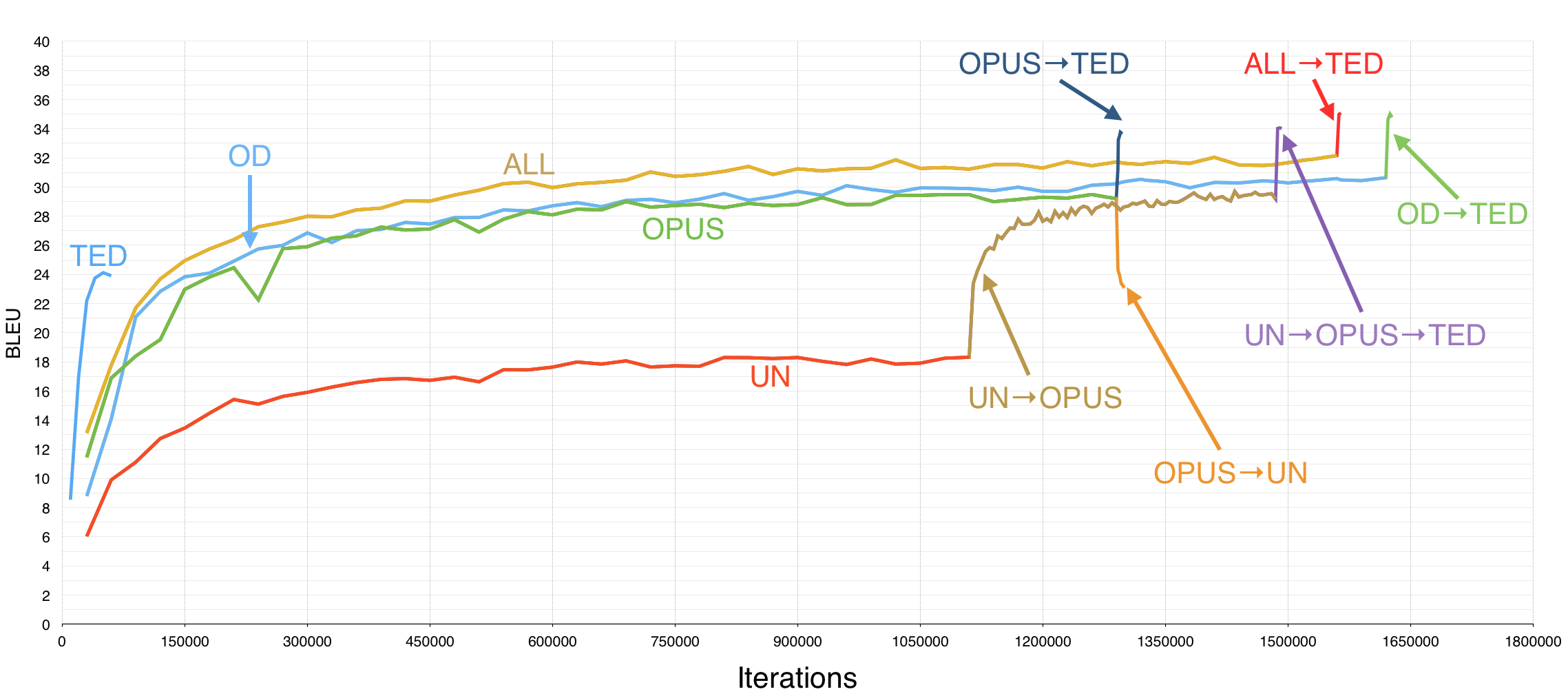}
	\caption{Arabic-English system development life line evaluated on development set tst-11 and tst-12. Here, \texttt{ALL} refers to \texttt{UN+OPUS+TED}, and \texttt{OD} refers to \texttt{UN+OPUS}}
	\label{fig:ar-curves}
\end{figure*}

\begin{figure*}[t]
	\centering
	\includegraphics[width=\textwidth]{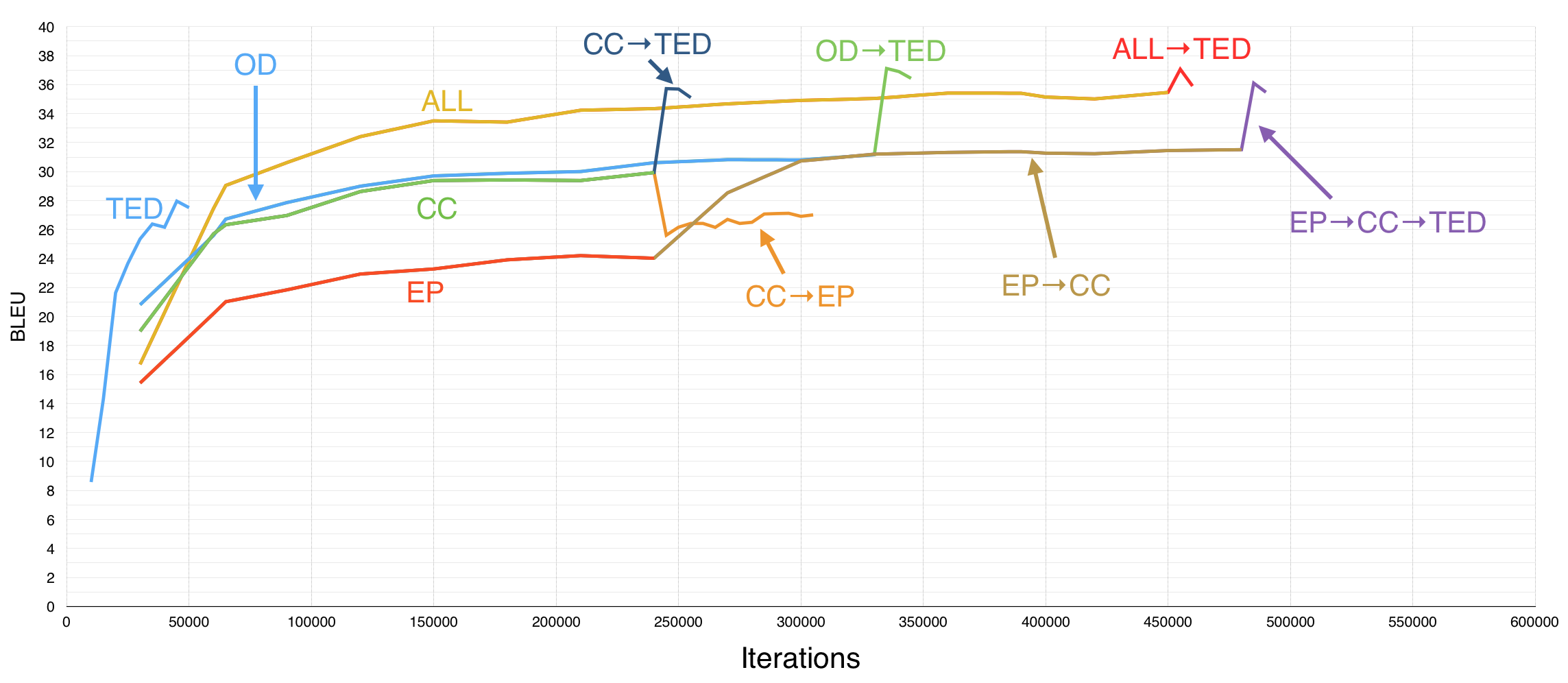}
	\caption{German-English system development life line evaluated on development set tst-11 and tst-12. Here, \texttt{ALL} refers to \texttt{EP+CC+TED}, and \texttt{OD} refers to \texttt{EP+CC}}
	\label{fig:de-curves}
\end{figure*}

In this section, we empirically compare several approaches to combine in- and out-of-domain data to train an NMT system. Figure \ref{fig:ar-curves} and Figure \ref{fig:de-curves} show the learning curve on development sets using various approaches mentioned in this work. We will go through them individually later in this section.



\subsection{Individual Systems}

We trained systems on each domain individually (for 10 epochs)\footnote{For German-English, we ran the models until they converged because the training data is much smaller compared to Arabic-English direction} and chose the best model using the development set. We tested every model on the in-domain testsets. Table \ref{tab:baseline} shows the results. On Arabic-English, the system trained on the out-of-domain data OPUS performed the best. This is due to the large size of the corpus and its spoken nature which makes it close to TED in style and genre. 
However, despite the large size of UN, the system trained using UN performed poorly. The reason is the difference in genre of UN from the TED corpus where 
the former consists of United Nations proceedings and 
the latter is based on talks. 

For German-English, the systems built using out-of-domain corpora performed better than the in-domain corpus. 
The CC corpus appeared to be very close to the TED domain.
The system trained on it performed even better than the in-domain system by an average of 2 BLEU points.

\begin{table}
\centering
\begin{tabular}{lrrrr}
\toprule
\multicolumn{4}{c}{\bf Arabic-English} \\
& TED & UN & OPUS & \\
\midrule
tst13 & 23.6 & 22.4 & {\bf 32.2}   \\
tst14 & 20.5 & 17.8 & {\bf 27.3}   \\
avg. & 22.1 & 20.1 & {\bf 29.7}   \\
\midrule
\multicolumn{4}{c}{\bf German-English} \\
& TED  & CC & EP \\
\midrule
tst13 & 29.5 & {\bf 29.8} & 29.1 \\
tst14 & 23.3  & {\bf 25.7} & 25.1 \\
avg. & 26.4 & {\bf 27.7} & 27.1 \\
\bottomrule
\end{tabular}
\caption{\label{tab:baseline} Individual domain models evaluated on TED testsets}
\end{table}

\subsection{Concatenation and Fine-tuning}
Next we evaluated how the models performed when trained on concatenated data. We mainly tried two variations: 
i) concatenating all the available data (\emph{ALL}) ii) combine only the available out-of-domain data (\emph{OD}) and later fine-tune the model on the in-domain data. Table \ref{tab:concatenate} shows the results. The fine-tuned system outperformed a full concatenated system by 1.8 and 2.1 average BLEU points in Arabic-English and German-English systems respectively.

Looking at the development life line of these systems (Figures \ref{fig:ar-curves}, \ref{fig:de-curves}), since \emph{ALL} has seen all of the data, it is better than \emph{OD } till the point \emph{OD} is fine-tuned on the in-domain corpus. Interestingly, at that point \emph{ALL} and \emph{OD}$\rightarrow$TED have seen the same amount of data but the parameters of the latter model are fine-tuned towards the in-domain data. This gives it 
average improvements of up to 2 BLEU points over \emph{ALL}.   

The \emph{ALL} system does not give any explicit weight to any domain \footnote{other than the data size itself} 
during training. In order to revive the in-domain data, we fine-tuned it on the in-domain data. We achieved comparable results to that of the OD$\rightarrow$TED model which means that one can adapt an already trained model on all the available data to a specific domain by fine tuning it on the domain of interest. This can be helpful in cases where in-domain data is not known beforehand. 


\begin{table}
\centering
\begin{tabular}{lccc|c}
\toprule
\multicolumn{5}{c}{\bf Arabic-English} \\
& TED & ALL & OD$\rightarrow$TED & ALL$\rightarrow$TED \\
\midrule
tst13 & 23.6 & 36.1 & 37.9 & 38.0 \\
tst14 & 20.5 & 30.2 & 32.1 & 32.2 \\
avg. & 22.1 & 33.2 & 35.0 & 35.1 \\
\midrule
\multicolumn{5}{c}{\bf German-English} \\
& TED & ALL & OD$\rightarrow$TED & ALL$\rightarrow$TED\\
\midrule
tst13 & 29.5 & 35.7 &  38.1 & 38.1 \\
tst14 & 23.3 & 30.8 &  32.8 & 32.9 \\
avg. & 28.0 & 33.3 &  35.4 & 35.5 \\
\bottomrule
\end{tabular}
\caption{\label{tab:concatenate} Comparing results of systems built on a concatenation of the data. OD represents a concatenation of the out-of-domain corpora and ALL represents a concatenation of OD and the in-domain data. $\rightarrow$ sign means fine-tuning}
\end{table}

\subsection{Model Stacking}

Previously we concatenated all out-of-domain data and fine-tuned it with the in-domain TED corpus. 
In this approach,
we picked one out-of-domain corpus at a time, trained a model and fine-tuned it with the other available domain. We repeated this process till all out-of-domain data had been used. In the last step, we fine-tuned the model on the in-domain data. 
Since we have a number of out-of-domain corpora available, we experimented with using them in different permutations for training and analyzed their effect on the development sets. Figure \ref{fig:ar-curves} and Figure \ref{fig:de-curves} show the results. It is interesting to see that the order of stacking has a significant effect 
on achieving a high quality system. The best combination for 
the Arabic-English language pair started with the UN data, fine-tuned on OPUS and then fine-tuned on TED. 
When we started with OPUS and fine-tuned the model on UN, the results dropped drastically as shown in Figure \ref{fig:ar-curves} (see OPUS$\rightarrow$UN). The model started forgetting the previously used data and focused on the newly provided data which is very distant from the in-domain data. We saw similar trends in the case of German-English language pair where CC$\rightarrow$EP dropped the performance drastically. We did not fine-tune CC$\rightarrow$EP and OPUS$\rightarrow$UN on TED since there was no better model to fine-tune than to completely ignore the second corpus i.e. UN and EP for Arabic and German respectively and fine-tune OPUS and CC on TED. The results of OPUS$\rightarrow$TED and CC$\rightarrow$TED are shown in Figures.

Comparing the OPUS$\rightarrow$TED system with the UN$\rightarrow$OPUS$\rightarrow$TED system, 
the result of OPUS$\rightarrow$TED are lowered by 0.62 BLEU points from the UN$\rightarrow$OPUS$\rightarrow$TED system. 
Similarly, we saw a drop of 0.4 BLEU points for German-English language pair when we did not use EP and directly fine-tuned CC on TED. 
There are two ways to look at these results,
considering
quality vs. time: i) by using UN and EP in model stacking, 
the
model learned to remember only those parts of the data that are beneficial 
for achieving
better translation quality on the in-domain development sets. Thus using them as part of the training pipeline is helpful 
for building a
better system. ii) training on UN and EP is expensive. Dropping them from the pipeline significantly reduced the training time and resulted in a loss of 
0.62 and 0.4 BLEU points only. 

To summarize, model stacking performs best when it starts from the 
domain furthest 
from the in-domain data. In the following, we compare it with the data concatenation approach. 


\subsection{Stacking versus Concatenation}

We compared model stacking with different forms of concatenation. In terms of data usage, all models are exposed to identical data. Table \ref{tab:stackvscat} shows the results. The best systems are achieved using a concatenation of all of the out-of-domain data for initial model training and then fine-tuning the trained model on the in-domain data. The concatenated system \emph{ALL} performed the lowest among all.  

\emph{ALL} learned a generic model from all the available data without giving explicit weight to any particular domain whereas model stacking resulted in a specialized system for the in-domain data. In order to confirm the generalization ability of \emph{ALL} vs. model stacking, we tested them on a new domain, News. \emph{ALL} performed 4 BLEU points better than model stacking in translating the news NIST MT04 testset. 
This concludes that a concatenation system is not an optimum solution for one particular domain but is robust enough to perform well in new testing conditions.


\begin{table}
\centering
\begin{tabular}{lccc}
\toprule
\multicolumn{4}{c}{\bf Arabic-English} \\
&  ALL & OD$\rightarrow$TED & UN$\rightarrow$OPUS$\rightarrow$TED \\
\midrule
tst13 & 36.1 & 37.9 & 36.8\\
tst14 & 30.2 & 32.1  & 31.2\\
avg. & 33.2 & 35.0  & 34.0 \\
\midrule
\multicolumn{4}{c}{\bf German-English} \\
& ALL & OD$\rightarrow$TED & EP$\rightarrow$CC$\rightarrow$TED \\
\midrule
tst13 & 35.7 & 38.1 &  36.8 \\
tst14 & 30.8 & 32.8 &  31.7 \\
avg. & 33.3 & 35.4 &  34.3 \\
\bottomrule
\end{tabular}
\caption{\label{tab:stackvscat} Stacking versus concatenation}
\end{table}

\subsection{Data Selection}
Since training on large out-of-domain data is time inefficient, we selected a small portion of 
out-of-domain data that is closer to the in-domain data. For Arabic-English, we selected 3\% and 5\% from the UN and OPUS data respectively which constitutes roughly ~2M sentences. For German-English, we selected 20\% from a concatenation of EP and CC, which roughly constitutes 1M training sentences.\footnote{These data-selection percentages have been previously found to be optimal when training phrase-based systems using the same data. For example see \cite{sajjad-etal:iwslt13}.}
 
We concatenated the selected data and the in-domain data to train an NMT system.
Table \ref{tab:selection} presents the results. 
The selected system is worse than the \emph{ALL} system. This is 
in
contrary to the results mentioned in the literature 
on
phrase-based machine translation where data selection on UN improves translation quality \cite{sajjad-etal:iwslt13}. This shows that NMT is not as sensitive as phrase-based to the presence of the out-of-domain data. 

Data selection comes with a cost of reduced translation quality. 
However, the selected system is better than all individual systems shown in Table \ref{tab:baseline}. Each of these out-of-domain systems take more time to train than a selected system. For example, compared to individual UN system, the selected system took approximately 1/10th of the time to train.
One can look at data selected system as a decent trade-off between training time and translation quality.

\begin{table}
\centering
\begin{tabular}{lcc|cc}
\toprule
&\multicolumn{2}{c}{\bf Arabic-English}&\multicolumn{2}{c}{\bf German-English} \\
&  ALL & Selected & ALL & Selected \\
\midrule
tst13 & 36.1 & 32.7   & 35.7 & 34.1 \\
tst14 & 30.2 & 27.8   & 30.8 & 29.9 \\
avg. & 33.2 & 30.3    & 33.3 & 32.0 \\
\bottomrule
\end{tabular}
\caption{\label{tab:selection} Results of systems trained on a concatenation of selected data and on a concatenation of all available data}
\end{table}


\subsection{Multi-domain Ensemble}

We took 
the 
best model 
for every domain according to the average BLEU on the development sets and ensembled them during decoding. 
For weighted ensemble, we did a grid search and selected the weights using the development set.
Table \ref{tab:ensemble} presents the results of
an ensemble on the
Arabic-English language pair and compares them with the individual best model, OPUS, 
and a model built on \emph{ALL}. As expected, balanced ensemble (\emph{ENS$_b$}) dropped results 
compared to 
the best individual model. Since 
the
domains are very distant, giving equal weights to them hurts the overall performance. The weighted ensemble (\emph{ENS$_w$}) improved from the best individual model by 1.8 BLEU points but is still lower than the concatenated system by 1.7 BLEU points. The weighted ensemble approach is beneficial when individual domain specific models are already available for testing. Decoding with multiple models is more efficient compared to training a system from scratch on a concatenation of the entire data.

\begin{table}
\centering
\begin{tabular}{lcc|cc}
\toprule
\multicolumn{5}{c}{\bf Arabic-English} \\
& OPUS & ALL & ENS$_b$ & ENS$_w$ \\
\midrule
tst13 & 32.2 & 36.1 & 31.9 & 34.3\\
tst14 & 27.3 & 30.2 & 25.8  & 28.6\\
avg. & 29.7 & 33.2 & 28.9 & 31.5 \\
\bottomrule
\end{tabular}
\caption{\label{tab:ensemble} Comparing results of balanced ensemble (ENS$_b$) and weighted ensemble (ENS$_w$) with the best individual model and the concatenated model}
\end{table}

\subsection{Discussion}

The concatenation system showed robust behavior in translating new domains. \cite{KobusCS16} proposed a domain aware concatenated system by introducing domain tags for every domain. We trained a system using their approach and compared the results with simple concatenated system. The domain aware system performed slightly better than the concatenated system (up to 0.3 BLEU points) when tested on the in-domain TED development sets. However, domain tags bring a limitation to the model since it can only be tested on the domains it is trained on. Testing on an unknown domain would first require to find its closest domain from the set of domains the model is trained on. The system can then use that tag to translate unknown domain sentences. 


\section{Conclusion}
\label{sec:conclusion}
We explored several approaches to train a neural machine translation system under multi-domain conditions and evaluated them based on three metrics: translation quality, training time and robustness. Our results showed that an optimum in-domain system can be built using a concatenation of the out-of-domain data and then fine-tuning it on the in-domain data. A system built on the concatenated data resulted in a generic system that is robust to new domains. Model stacking is sensitive to the order of domains it is trained on. Data selection and weighted ensemble resulted in a less 
optimal
solution. 
The
former is efficient to train in a short time and 
the
latter is useful when different individual models are available for testing. It provides a mix of all
domains
without retraining or fine-tuning the system. 

\section{Acknowledgments}
The research presented in this paper is partially conducted as part of the European Union’s Horizon 2020
research and innovation programme under grant
agreement 644333 (SUMMA).

\bibliographystyle{IEEEtran}
\bibliography{emnlp2017}

\end{document}